\pdfoutput=1

\documentclass[11pt]{article}

\usepackage{acl}

\usepackage{times}
\usepackage{latexsym}
\usepackage{float}

\usepackage[T1]{fontenc}

\usepackage[utf8]{inputenc}

\usepackage{microtype}
\usepackage{amsmath, amssymb}
\usepackage{booktabs}
\usepackage{graphicx}
\usepackage{multirow}

\usepackage{bm}
\usepackage{algorithm}
\usepackage{pifont}
\usepackage{subfigure}
\usepackage{xspace}

\newcommand{\modelname}{\textsc{Esd}\xspace}

\newcommand{\bl}[1]{\textbf{#1}}
\title{An Enhanced Span-based Decomposition Method for Few-Shot \\ Sequence Labeling}

\author{
Peiyi Wang$^1$$\footnotemark[1]$ $\footnotemark[3]$, \ Runxin Xu$^1$$\footnotemark[1]$, \  Tianyu Liu$^2$, \  Qingyu Zhou$^2$, \\
\textbf{Yunbo Cao$^2$, \  Baobao Chang$^1$, \  Zhifang Sui$^1$$\footnotemark[2]$}
\\ 
$^1$ Key Laboratory of Computational Linguistics, Peking University, MOE, China \\
$^2$ Tencent Cloud Xiaowei \\
 \texttt{\{wangpeiyi9979, runxinxu\}@gmail.com} \\
 \texttt{\{rogertyliu, qingyuzhou, yunbocao\}@tencent.com} \\
 \texttt{\{chbb, szf\}@pku.edu.cn}
}

\begin{document}

\maketitle
\renewcommand{\thefootnote}{\fnsymbol{footnote}}
\begin{abstract}
Few-Shot Sequence Labeling (FSSL) is a canonical paradigm for the tagging models, e.g., named entity recognition and slot filling, to generalize on an emerging, resource-scarce domain.
Recently, the metric-based meta-learning framework has been recognized as a promising approach for FSSL.
However, most prior works assign a label to each token based on the token-level similarities, which ignores the integrality of named entities or slots.
To this end, in this paper, we propose \textbf{\modelname}, an \underline{\textbf{E}}nhanced \underline{\textbf{S}}pan-based \underline{\textbf{D}}ecomposition method for FSSL.
\modelname formulates FSSL as a span-level matching problem between test query and supporting instances.
Specifically, \modelname decomposes the span matching problem into a series of span-level procedures, mainly including enhanced span representation, class prototype aggregation and span conflicts resolution.
Extensive experiments show that \modelname achieves the new state-of-the-art results on two popular FSSL benchmarks, FewNERD and SNIPS, and is proven to be more robust in the nested and noisy tagging scenarios. 
Our code is available at \url{https://github.com/Wangpeiyi9979/ESD}.

\end{abstract}
\footnotetext[1]{Equal contribution.}
\footnotetext[2]{Corresponding author.}
\footnotetext[3]{Contribution during internship in Tencent.}

\renewcommand{\thefootnote}{\arabic{footnote}}

\section{Introduction}

\begin{figure}[t]
    \centering
    \includegraphics[width=0.9\linewidth]{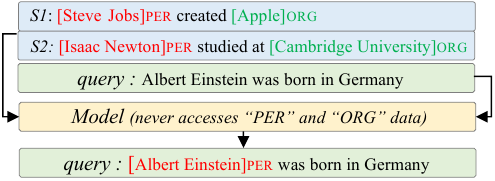}
    \caption{
    A $2$-way $2$-shot few-shot NER task. The model needs to learn new entities with few examples.}
    \label{fig:intro}
\end{figure}

Many natural language understanding tasks can be formulated as sequence labeling tasks,
such as \textit{Named Entity Recognition} (NER) and \textit{Slot Filling} (SF) tasks.
Most prior works on sequence labeling follow the supervised learning paradigm, which requires large-scale annotated data and is limited to pre-defined classes.
In order to generalize on the emerging, resource-scare domains, Few-Shot Sequence Labeling (FSSL) has been proposed \cite{fewshothou, NNshot}.
In FSSL, the model (typically trained on the source domain data) needs to solve the $N$-way ($N$ unseen classes) $K$-shot (only $K$ annotated examples for each class) task in the target domain.
Figure \ref{fig:intro} shows a $2$-way $2$-shot target domain few-shot NER task,
where \textit{`PER' }and \textit{`ORG'} are $2$ unseen entity types, and in the training set $\mathcal{S}=\{S_1, S_2\}$, both of them have only $2$ annotated entities.
The tagging models should annotate \textit{`Albert Einstein'} in the test sentence $q$ as a \textit{`PER'} according to $\mathcal{S}$.

Recently, \textit{metric-based meta-learning (MBML) methods} have become the mainstream and state-of-the-art methods in FSSL \cite{fewshothou,FewNERD}, which train the models on the tasks sampled from the source domain sentences in order to mimic and solve the target task.
In each task of training and testing, they make the prediction through modeling the similarity between the training set (support set) and the test sentence (query).
Specifically, previous MBML methods \cite{FewNERD, fewshothou, NNshot} mainly formulate FSSL as a token-level matching problem, which assigns a label to each token based on the token-level similarities. For example, the token \textit{`Albert'} would be labeled as \textit{`B-PER'} due to its resemblance to \textit{`Steve'} and \textit{`Isaac'}.

However, for the MBML methods, 
selecting the appropriate metric granularity would be fundamental to the success.
We argue that prior works that focus on modeling the token-level similarity are sub-optimal in FSSL:
1) they ignore the integrality of named entities or dialog slots that are composed of a text span instead of a single token.
2) in FSSL, the conditional random fields (CRF) is an important component for the token-level sequence labeling models \cite{NNshot, fewshothou}, but the transition probability between classes in the target domain task can not be sufficiently learned with very few examples \cite{span-naacl}. The previous methods resort to estimate the values with the abundant source domain data, which may suffer from domain shift problem \cite{fewshothou}.
To overcome these drawbacks of the token-level models,
in this paper, we propose \modelname, an \textbf{E}nhanced \textbf{S}pan-based \textbf{D}ecomposition model that formulates FSSL as a span-level matching problem between test query and support set instances.

Specifically, \modelname decomposes the span matching problem into three main subsequent span-level procedures.
1) \textbf{Span representation}. 
We find the span representation can be enhanced by information from other spans in the same sentence and the interactions between query and support set.
Thus we propose a span-enhancing module to reinforce the span representation by intra-span and inter-span interactions.
2) \textbf{Span prototype aggregation}.
MBML methods usually aggregate the span vectors that belongs to the same class in the support set to form the class prototype representation. 
Among all class prototypes, the O-type serves as negative examples and covers all the miscellaneous spans that are not entities, which poses new challenges to identify the entity boundaries.
To this end, we propose a span-prototypical module to divide O-type spans into $3$ sub-types to distinguish the miscellaneous semantics by their relative position with recognized entities, together with a dynamically aggregation mechanism to form the specific prototype representation for each query span.
3) \textbf{Span conflicts resolution}.
In the span-level matching paradigm, the predicted spans may conflict with each other. 
For example, a model may annotate both ``Albert Einstein'' and ``Albert'' as \textit{``PER''}.
Therefore, we propose a span refining module that incorporates the Soft-NMS \cite{soft-NMS,SoftNMS-NER} algorithm into the beam search to alleviate this conflict problem.
With Beam Soft-NMS, \modelname can also handle nested tagging cases without any extra training, which previous methods are incapable of.                

We summarize our contribution as follows: 
(1) We propose \modelname, an enhanced span-based decomposition model, which formulates FSSL as a span-level matching problem. 
(2) We decompose the span matching problem into 3 main subsequent procedures, which firstly produce enhanced span representation, then distinguish miscellaneous semantics of O-types, and achieve the specific prototype representation for each query, and finally resolve span conflicts . 
(3) Extensive experiments show that \modelname achieves new state-of-the-art performance on both few-shot NER and slot filling benchmarks and that ESD is more robust than other methods in nested and noisy scenarios.

\begin{figure*}[h]
    \centering
    \includegraphics[width=\linewidth]{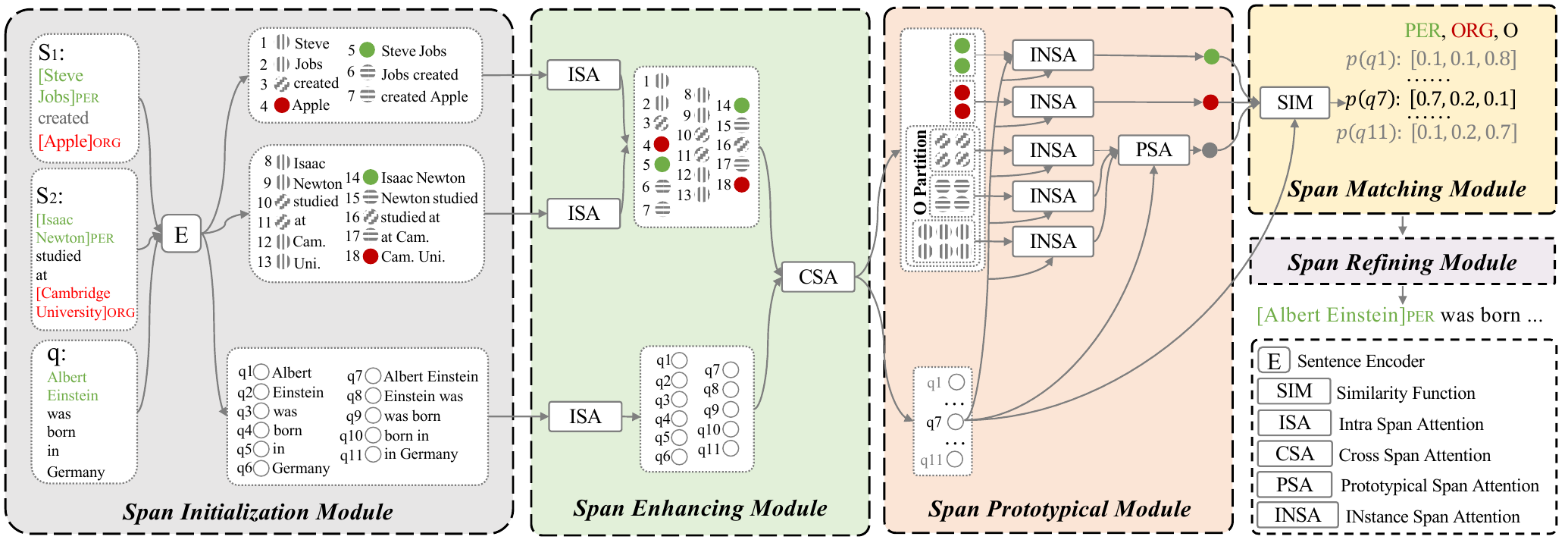}
    \caption{The architecture ($5$ span modules) of \modelname with a $2$-way ({`PER'} and {`ORG'}) $2$-shot input task. We only list spans with lengths less than $2$ for clarity. \modelname assigns label {`PER'} to the span {``Albert Einstein''} and {`O'} to the other spans in query $q$ based on the support set $\{S_1, S_2\}$. 
}
    \label{fig:overview}
\end{figure*}

\section{Related Work}
\subsection{Few-Shot Learning}
Few-Shot Learning (FSL) aims to enable the model to solve the target domain task with a very small training set $D_{train}$ \cite{fsl}.
In FSL, people usually consider the $N$-way $K$-shot task, where the $D_{train}$ has $N$ classes, and each class has only $K$ annotated examples ($K$ is very small, e.g., $5$).
Training a model only based on $D_{train}$ from scratch will inevitably lead to over-fitting.
Therefore, researchers usually introduce the source domain annotated data to help train models, e.g., Few-Shot Relation Classification (FSRC) \cite{fewrel}, Few-Shot Text Classification (FSTC) \cite{fewtc} and Few-Shot Event Classification (FSEC) \cite{wang2021behind}.
The source domain data do not contain examples belonging to classes in $D_{train}$, and thus the FSL setting can be guaranteed.
Specifically, FSSL tasks in our paper also have the source domain data.

\subsection{Metric-Based Meta-Learning}
Meta-learning \cite{meta-learning} is a popular method to deal with Few-Shot Learning (FSL).
Meta-learning constructs a series of tasks sampled from the source domain data to mimic the target domain task, and trains models across these sampled tasks.
Each task contains a training set (support set) and a test instance (query).
The core idea of meta-learning is to help model learn the ability to quickly adapt to new tasks, i.e., learn to learn \cite{l2l}. Meta-learning can be combined with the metric-learning \cite{metric-learning-survey} (metric-based meta-learning), which makes predictions based on the similarity of the support set and the query.
For example, Prototypical Network \cite{Proto} first learns prototype vectors from a few examples in the support set for classes, and further uses prototype vectors for query prediction. Specifically, \modelname (our model) follows this metric-based meta-learning paradigm.

\subsection{Few-Shot Sequence Labeling}
Recently, few shot sequence labeling (FSSL) has been widely explored by researchers.
For example, \cite{fritzler2019few} leverages the prototypical network \cite{Proto} in the few-shot NER.
\cite{NNshot} further proposes a cheap but effective method to capture the label dependencies between entity tags without expensive CRF training.
\cite{wang2021meta} utilizes a large unlabelled dataset and proposes a distillation method.
\cite{cui-etal-2021-template} introduces a prompt method to tap the potential of BART \cite{lewis2020bart}.
\cite{fewshothou} extends the TapNet \cite{yoon2019tapnet} and proposes a collapsed CRF for few-shot slot filling.
\cite{mrc-few-slot} and \cite{GSL} formulate FSSL as a machine comprehension problem and a generation problem, respectively.
\cite{span-naacl} proposed to retrieve the most similar exemplars in the support set for span-level prediction. 
Although their methods also involve span matching, their main focus is on the retrieval-augmented training. 
Our work differs from \cite{span-naacl} in that we propose an effective actionable pipeline to get enhanced span representation, handle miscellaneous semantics of O-types and resolve the potential span conflicts in both non-nested and nested situations, where the last two modules are essential to align the candidate spans with class prototypes in the support set but missing in \cite{span-naacl}.
Besides, our enhanced span representation greatly improves the batched softmax objective of \cite{span-naacl} by inter- and intra-span interactions.

\section{Task Formulation}
We define a sentence as $\mathbf{x}$ and its label as $\mathbf{y} = \{(s_i, y_i )\}_{i=1}^{M}$, where $s_i$ is a span of $\mathbf{x}$ (e.g., \textit{`Steve Jobs’}), $y_i$ is the label of $s_i$ ( e.g., \textit{`PER'}) and $M$ is the number of spans in the $\mathbf{x}$.
Following the previous FSSL setting \cite{fewshothou, FewNERD}, we have data in source domain $\mathcal{D}_{source}$ and target domain $\mathcal{D}_{target}$, and models are evaluated on tasks from $\mathcal{D}_{target}$.
Meta-learning based FSSL has two stages, meta-training and meta-testing. 
In the meta-training stage, the model is trained on tasks sampled from $\mathcal{D}_{source}$ to mimic the test situation, and in the meta-testing stage, the model is evaluated on test tasks.
A task is defined as $\mathcal{T} = \{\mathcal{S}, q\}$, consisting of a support set $\mathcal{S} = \{(\mathbf{x}_i, \mathbf{y}_i)\}_{i=1}^{I}$,  and a query sentence $q = \mathbf{x}$. 
$\mathcal{S}$ includes $N$ types of entities or slots ($N$-way), and each type has $K$ annotated examples ($K$-shot). 
For spans that do not belong to the $N$ types, e.g., \textit{`studied at'} in Figure \ref{fig:intro}, we set their label to O.
The types except O in each test task are guaranteed to not exist in the training tasks.
Given a task $\mathcal{T} = \{\mathcal{S}, q\}$,  the model needs to assign a label to each span in the query sentence $q$ based on $\mathcal{S}$.

\section{Methodology}
Our \modelname formulates FSSL as a span-level matching problem and decomposes it into a series of span-related procedures for a better span matching.
Figure \ref{fig:overview} illustrates the architecture of \modelname.

\subsection{Span Initialization Module}
Given a task $\mathcal{T} = \{\mathcal{S}, q\}$, we use BERT \cite{Bert} to encode sentences in $\mathcal{S}$ and $q$, and utilize the output of the last layer to represent tokens in the sentence.
Therefore, for a sentence with $N$ tokens $\mathbf{x} = (x_1, x_2, ..., x_N)$, we can achieve representations $(\mathbf{h_1}, \mathbf{h_2}, ..., \mathbf{h_N})$, where $\mathbf{h_i} \in \mathbb{R}^{d_w}$ is the hidden state corresponding to the token $x_i$.
Then, for a span $s = (l, r)$, where $l$ and $r$ are the start index and end index of span $s$ in the sentence $\mathbf{x}$, we obtain its initial representation $\mathbf{s}_{(l, r)}= [\mathbf{h_l}; \mathbf{h_r}]\mathbf{W_{s}}$.\footnote{We omit the bias term in this paper for clarity.}
We enumerate spans in the sentence with a maximum length of $L$.

\subsection{Span Enhancing Module}
\subsubsection{Intra Span Attention}
Intuitively, the meaning of specific spans can be inferred from other spans in the same sentence.
We thus design an Intra Span Attention (ISA) mechanism.
Given all the span representations of a sentence $\mathbf{S} \in \mathbb{R}^{B \times d}$ ($B$ is the number of spans).
We denote the $i$-th row of $\mathbf{S}$ as $\mathbf{s_i}$, which represents the $i$-th span in the sentence. 
For $\mathbf{s_i}$, we first get its ISA span representation 
$\mathbf{\bar{s}_i}=\sum_{j=1}^B\mathbf{\alpha_i}^j\mathbf{s_j}$,
where $\mathbf{\alpha_{i}} = {\rm softmax(\mathbf{s_i}\mathbf{S^T})}$. 
For clarity, we denote this attention aggregation operation as $\phi$, i.e., 
\begin{equation}
    \mathbf{\bar{s}_i}=\phi(\mathbf{{s}_i}, \mathbf{S})
\end{equation}
then, a Feed Forward Neural Networks (FFN) \cite{Transformer} with residual connection \cite{ResNet} and layer normalization \cite{LayerNorm} are used to get the final ISA enhanced feature $\mathbf{\tilde{s}_i}$,
\begin{gather}
    \mathbf{\tilde{s}_i} = {\rm LayerNorm}(\mathbf{s_i} + {\rm FFN_{isa}}(\mathbf{\bar{s}_i})) \\
    {\rm FFN_{isa}}(\mathbf{\bar{s}_i}) = {\rm GELU}(\mathbf{\bar{s}_i}\mathbf{W^1_{isa}})\mathbf{W^2_{isa}}
\end{gather}

\subsubsection{Cross Span Attention}
After ISA, to accommodate the span matching between the test query and supporting sentences, and facilitate the inter-span interaction, we propose Cross Span Attention (CSA) to enhance query spans $\mathbf{\tilde{Q}} \in \mathbb{R}^{B_q \times d}$ with the support set spans $\{\mathbf{\tilde{S}_i} \in \mathbb{R}^{B_i \times d}; i=1, ..., I\}$, and vice versa.
We first concatenate all span representations in the support set into one matrix $\mathbf{\tilde{S}} = [\mathbf{\tilde{S}_1}, \mathbf{\tilde{S}_2}, ... , \mathbf{\tilde{S}_I}] \in \mathbb{R}^{B_s \times d}$.
We denote the $n$-th row of $\mathbf{\tilde{S}}$ as $\mathbf{\tilde{s}_n}$ and the $m$-th row of $\mathbf{\tilde{Q}}$ as $\mathbf{\tilde{q}_m}$, 
and obtain their CSA span representations
$\mathbf{\hat{s}_n}=\phi(\mathbf{\tilde{s}_n},\mathbf{\tilde{Q}})$ and 
$\mathbf{\hat{q}_m}=\phi(\mathbf{\tilde{q}_m},\mathbf{\tilde{S}})$.
Then, we get the final CSA enhanced representation $\mathbf{\check{s}_n}$ and $\mathbf{\check{q}_m}$ as follows,
\begin{gather}
	\mathbf{\check{s}_n} = {\rm LayerNorm}(\mathbf{\tilde{s}_n} + {\rm FFN_{csa}}(\mathbf{\hat{s}_n})) \\
	\mathbf{\check{q}_m} = {\rm LayerNorm}(\mathbf{\tilde{q}_m} + {\rm FFN_{csa}}(\mathbf{\hat{q}_m})) \\
	 {\rm FFN_{csa}}(\mathbf{x}) = \mathrm{GELU}(\mathbf{x}\mathbf{W_{csa}^1})\mathbf{W_{csa}^2}
\end{gather}

\subsection{Span Prototypical Module}
\subsubsection{Instance Span Attention}
Since different support set spans play different roles for a query span, we propose multi-INstance Span Attention (INSA) to get class representations.
For the $i$-th class that contains $K$ annotated spans with enhanced representations $\mathbf{\check{S}_i}=[ \mathbf{\check{s}_i}^1,...,\mathbf{\check{s}_i}^K]$ in the support set,
given a query span $\mathbf{\check{q}_{m}}$,  INSA gets the corresponding prototypical representation $\mathbf{z^i_m}=\phi(\mathbf{\check{q}_{m}}, \mathbf{\check{S}_i})$.

\subsubsection{O Partition and Prototypical Span Attention}
The O-type spans have huge quantities and miscellaneous semantics, which is hard to be represented by only one prototypical vector. In the span-based framework, considering the boundary information is essential for a span, we divide the O-type spans into $3$ sub-classes according to their boundary, to alleviate their miscellaneous semantics problem.
Specifically, given a sentence with $I$ annotated spans $\{(l_i, r_i) \}_{i=1}^I$, where $l_i$ and $r_i$ are the left and right boundary of the $i$-th annotated span.
For each of the other spans $(l_o, r_o)$ , we assign it a sub-class O$_{sub}$ as follows,
\begin{equation}
    \begin{aligned}
        O_{sub} =\left\{
        \begin{aligned}
         &O_{1}, &  \forall i, s.t. \ r_o < l_i \lor l_o > r_i \\
         &O_{2}, &  \exists i, s.t. \ l_o \ge l_i \land r_o \le r_i \\\
         &O_{3}, & {\rm Others} \ \ \ \ \ \ \ \ \ \ \ \ \ \ \ \ \\
        \end{aligned}
        \right. 
    \end{aligned}
\end{equation}
where $O_1$ denotes the span that does not overlap with any entities or slots in the sentence, e.g., \textit{``study at''} in $S_2$ of Figure \ref{fig:overview},
and $O_2$ represents the span that is the sub-span of an entity or slot, e.g., \textit{``Isaac''} in $S_2$ of Figure  \ref{fig:overview}. 
After O Partition, we get the prototypical representation of each O$_{sub}$ via INSA, thus for a query span $\mathbf{\check{q}_m}$, we have 3 sub-class representations $\mathbf{Z_m^o} = [\mathbf{z^{o_{1}}_m}, \mathbf{z^{o_{2}}_m}, \mathbf{z^{o_{3}}_m}]$ for the class O.
Then, we utilize Prototypical Span Attention (PSA) to achieve the final O representation 
$\mathbf{z^o_m} = \phi(\mathbf{\check{q}_{m}}, \mathbf{Z_m^o})$.
\begin{figure}[t]
    \centering
    \includegraphics[width=1\linewidth]{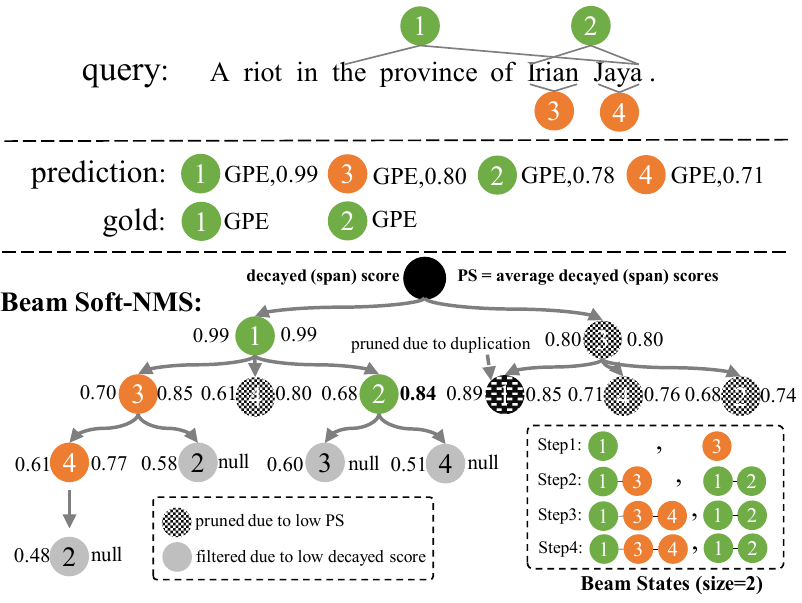}
    \caption{A demonstration for Beam Soft-NMS (BSNMS). 
    For clarity, we set the filter threshold $\delta$ to $0.6$, and suppose the span score is always decayed by the overlapped spans with a constant decayed score  $-0.1$ (Note that the example is simplified for demonstration). BSNMS filters false positive spans (colored in orange) in this demonstration. A more clear step by step demonstration is included in Appendix \ref{apx:bsnms}.
    }
    \label{fig:beam-softnms}
\end{figure}

\subsection{Span Matching Module}
Given a task $\mathcal{T} = (\mathcal{S}, q)$, for the $m$-th span in $q_m$, we achieve its enhanced representation $\mathbf{\check{q}_m}$ and corresponding prototypical vectors $\mathbf{Z_m} = (\mathbf{z^o_m}, \mathbf{z^1_m}, ..., \mathbf{z^N_m})$ through previous span modules.
Then, we predict $q_m$ as the type $z_k$ in the support set with probability,
\begin{equation}
    p(y_m=z_k|q_m) = \frac{exp(-L_2(\mathbf{\check{q}_m},  \mathbf{z_k^m}))}{\sum_{k'}exp(-L_2(\mathbf{\check{q}_m},  \mathbf{z_{k'}^m}))} 
\end{equation}
where $L_2$ is the euclidean distance.  
During training,  we use cross-entropy as our loss function,
\begin{equation}
    \mathcal{L} =-\frac{1}{B_q}\sum_{m=1}^{B_q}log \ p(y_m^*|q_m)
\end{equation}
where $y_m^*$ is the gold label of $q_m$ and $B_q$ is the number of spans in the query $q$.
\begin{table*}[t]
     \centering
    \scalebox{0.8}{
             \centering
            \begin{tabular}{lcccccccccc} \toprule
            \multirow{3}{*}{\textbf{Models}} & \multicolumn{5}{c}{\textbf{\textsc{Few-NERD (intra) }}} &  \multicolumn{5}{c}{\textbf{\textsc{Few-NERD (inter) }}}  \\
            \cmidrule(r){2-6} \cmidrule(r){7-11}
             & \multicolumn{2}{c}{$1\sim2$ shot} &  \multicolumn{2}{c}{$5\sim10$ shot} & \multirow{3}{*}{\textbf{Avg.}} &  \multicolumn{2}{c}{$1\sim2$ shot} &  \multicolumn{2}{c}{$5\sim10$ shot}  & \multirow{3}{*}{\textbf{Avg.}}     \\
             \cmidrule(r){2-3} \cmidrule(r){4-5}  \cmidrule(r){7-8} \cmidrule(r){9-10}
             & 5 way  & 10 way & 5 way & 10 way  & ~ & 5 way  & 10 way & 5 way & 10 way    \\

            \midrule
            ProtoBERT
            & 20.76\tiny{$\pm$0.84}   & 15.04\tiny{$\pm$0.44} & 42.54\tiny{$\pm$0.94} & 35.40\tiny{$\pm$0.13} & 28.44 
            
            &  38.83\tiny{$\pm$1.49} &  32.45\tiny{$\pm$0.79}  & 58.79\tiny{$\pm$0.44}  & 52.92\tiny{$\pm$0.37} & 45.75 \\
            NNShot
            & 25.78\tiny{$\pm$0.91}   & 18.27\tiny{$\pm$0.41} &  36.18\tiny{$\pm$0.79} & 27.38\tiny{$\pm$0.53} & 26.90
            
            & 47.24\tiny{$\pm$1.00} &  38.87\tiny{$\pm$0.21} & 55.64\tiny{$\pm$0.63}  & 49.57\tiny{$\pm$2.73}  & 47.83\\
            StructShot
            & 30.21\tiny{$\pm$0.90}   & 21.03\tiny{$\pm$1.13} &  38.00\tiny{$\pm$1.29} & 26.42\tiny{$\pm$0.60} & 28.92
            
            & 51.88\tiny{$\pm$0.69} &  43.34\tiny{$\pm$0.10}  & 57.32\tiny{$\pm$0.63}  & 49.57\tiny{$\pm$3.08} & 50.53\\
            \modelname (Ours)     
            & \bf{36.08\tiny{$\pm$1.6}}   & \bf{30.00\tiny{$\pm$0.70}} &  \bf{52.14\tiny{$\pm$1.5}} & \bf{42.15\tiny{$\pm$2.6}} & \bl{40.09} & 
            
            \bf{59.29\tiny{$\pm$1.25}} &  \bf{52.16\tiny{$\pm$0.79}}  & \bf{69.06\tiny{$\pm$0.80}}  & \bf{64.00\tiny{$\pm$0.43}} & \bl{61.13}
            \\ \bottomrule
            \end{tabular}
        }
       \caption{F1 scores with standard deviations on FewNERD. The best results are in \textbf{boldface}.}
      \label{tab:fewnerd}
\end{table*}

\begin{table*}[th]
\centering
\scalebox{0.85}{
    \begin{tabular}{p{0.2cm} lcccccccccccc}
    \toprule
    &  \textbf{Models} & We & Mu & Pl & Bo & Se & Re & Cr & \textbf{Avg.} \\ 
      \midrule
     \multirow{6}{*}{\rotatebox{90}{\textbf{\textsc{1-shot}}} }   
     & TransferBERT   & {55.82\tiny{$\pm$2.75}} & {38.01\tiny{$\pm$1.74}} & {45.65\tiny{$\pm$2.02}} & {31.63\tiny{$\pm$5.32}} & {21.96\tiny{$\pm$3.98}} & {41.79\tiny{$\pm$3.81}} & {38.53\tiny{$\pm$7.42}} & {39.06\tiny{$\pm$3.86}}  \\
     & MN+BERT  & {21.74\tiny{$\pm$4.60}} & {10.68\tiny{$\pm$1.07}} & {39.71\tiny{$\pm$1.81}} & {58.15\tiny{$\pm$0.68}} & {24.21\tiny{$\pm$1.20}} & {32.88\tiny{$\pm$0.64}} & {{69.66}\tiny{$\pm$1.68}} & {36.72\tiny{$\pm$1.67}}   \\
     & ProtoBERT   & {46.72\tiny{$\pm$1.03}} & {40.07\tiny{$\pm$0.48}} & {50.78\tiny{$\pm$2.09}} & {68.73\tiny{$\pm$1.87}} & {60.81\tiny{$\pm$1.70}} & {55.58\tiny{$\pm$3.56}} & {67.67\tiny{$\pm$1.16}} & {55.77\tiny{$\pm$1.70}}   \\
     &Ma2021 & - & - & - & - & - & - & - & 69.3\tiny{(unk)} \\
     & L-TapNet+CDT & {{71.53}\tiny{$\pm$4.04}} & {60.56}\tiny{$\pm$0.77} & {{66.27}\tiny{$\pm$2.71}} & {84.54\tiny{$\pm$1.08}} &  {76.27}\tiny{$\pm$1.72} & {{70.79}\tiny{$\pm$1.60}} & {62.89\tiny{$\pm$1.88}} & {{70.41}\tiny{$\pm$1.97}} \\ 
     & \modelname (Ours) & 78.25\tiny{$\pm$1.50} & {54.74\tiny{$\pm$1.02}} & 71.15\tiny{$\pm$1.55} & 71.45\tiny{$\pm$1.38} & 67.85\tiny{$\pm$0.75}& 71.52\tiny{$\pm$0.98}  &  78.14\tiny{$\pm$1.46} & \bf{70.44\tiny{$\pm$0.47}} \\
      \midrule
    
    \multirow{8}{*}{\rotatebox{90}{ \textbf{\textsc{5-shot}} }} 
     & TransferBERT
        & {59.41\tiny{$\pm$0.30}} & {42.00\tiny{$\pm$2.83}} & {46.07\tiny{$\pm$4.32}} & {20.74\tiny{$\pm$3.36}} & {28.20\tiny{$\pm$0.29}} & {67.75\tiny{$\pm$1.28}} & {58.61\tiny{$\pm$3.67}} & {46.11\tiny{$\pm$2.29}}  \\
     & MN+BERT 
        & {36.67\tiny{$\pm$3.64}} & {33.67\tiny{$\pm$6.12}} & {52.60\tiny{$\pm$2.84}} & {69.09\tiny{$\pm$2.36}} & {38.42\tiny{$\pm$4.06}} & {33.28\tiny{$\pm$2.99}} & {72.10\tiny{$\pm$1.48}} & {47.98\tiny{$\pm$3.36}}  \\
     & ProtoBERT  
        & {67.82\tiny{$\pm$4.11}} & {55.99\tiny{$\pm$2.24}} & {46.02\tiny{$\pm$3.19}} & {72.17\tiny{$\pm$1.75}} & {73.59\tiny{$\pm$1.60}} & {60.18\tiny{$\pm$6.96}} & {66.89\tiny{$\pm$2.88}} & {63.24\tiny{$\pm$3.25}}  \\
     & Retriever 
        & 82.95\tiny{(unk)}   & 61.74\tiny{(unk)}   & 71.75\tiny{(unk)}   & 81.65\tiny{(unk)}   & 73.10\tiny{(unk)}   & 79.54\tiny{(unk)}   & 51.35\tiny{(unk)}   & 71.72\tiny{(unk)}   \\
    & ConVEx  & 71.5\tiny{(unk)} & 77.6\tiny{(unk)} & 79.0\tiny{(unk)} & 84.5\tiny{(unk)} & 84.0\tiny{(unk)} & 73.8\tiny{(unk)} & 67.4\tiny{(unk)} & 76.8\tiny{(unk)} \\
     &Ma2021  & 89.39\tiny{(unk)} & 75.11\tiny{(unk)} & 77.18\tiny{(unk)} & 84.16\tiny{(unk)} & 73.53\tiny{(unk)} & 82.29\tiny{(unk)} & 72.51\tiny{(unk)} & 79.17\tiny{(unk)} \\
    & L-TapNet+CDT
        & {71.64\tiny{$\pm$3.62}} & {67.16\tiny{$\pm$2.97}} & {{75.88}\tiny{$\pm$1.51}} & 84.38\tiny{$\pm$2.81} & {{82.58}\tiny{$\pm$2.12}} & {{70.05}\tiny{$\pm$1.61}} & {{73.41}\tiny{$\pm$2.61}} & {{75.01}\tiny{$\pm$2.46}}  \\
     & \modelname (Ours)
        & 84.50\tiny{$\pm$1.06}& 66.61\tiny{$\pm$2.00}& 79.69\tiny{$\pm$1.35} & 82.57\tiny{$\pm$1.37} & 82.22\tiny{$\pm$0.81} & 80.44\tiny{$\pm$0.80}& 81.13\tiny{$\pm$1.84} & \bl{79.59\tiny{$\pm$0.39}}  \\
     \bottomrule
    \end{tabular}
    }
\caption{F1 scores with standard deviations on $7$ domains of SNIPS. The best results are in \textbf{boldface}. `unk' denotes methods that do not report deviations in their paper. 
Baselines of 1-shot and 5-shot settings are different since ConVEx and Retriever do not report the 1-shot results in their paper.} \label{tab:snips}
\end{table*}

\subsection{Span Refining Module in Inference} 
\label{sec:span_refining_module}
During inference, spans outputted by the matching module may have conflicts, we thus propose a refining module that incorporates the SoftNMS into beam search for the conflicts resolution. 
For the $m$-th span with the left index $l_m$ and the right index $r_m$ in the query, we obtain its prediction probability distribution $\mathbf{p_m}$, label $y_m = {\rm argmax}(\mathbf{p_m})$ and score $score_m={\rm max}(\mathbf{p_m})$.
Figure \ref{fig:beam-softnms} illustrates a simplified post-processing process.
In each step, we first expand all beam states (e.g., states 1-3 and 1-2 in step2 of Figure \ref{fig:beam-softnms}), and then prune new states according to the beam size.
Specifically, given a beam state $S$ containing spans $\{l_t, r_t, score_t, y_t\}_{t=1}^T$,
for each non-contained span $s_i=(l_i, r_i, score_i, y_i)$, we first calculate its decayed score $score_i^{decay}$,
\begin{gather}
     score_i^{decay} = score_i * u^{\eta} \\
     \eta = \sum_{t=1}^{T} \mathbb{I}({\rm IoU}(s_i, s_t) \ge k)
\end{gather}
where ${\rm IoU}(s_i, s_t) = \frac{|\{l_i, ..., r_i\} \cap \{l_t, ... , r_t \}|}{|\{l_i, ..., r_i\} \cup \{l_t, ... , r_t \}|}$ is the overlap ratio of two spans. The decay ratio $u$ and threshold $k$ are hyperparameters.
Then, we expand the beam state $S$ with the non-contained span $s_i$ if $score_i^{decay}>\delta$.
For example,  in the step2 of Figure \ref{fig:beam-softnms}, we expand the state 1-3 to 1-3-4, while the state 1-3-2 fails to be expanded since $score_2^{decay}=0.58 <= \delta$.
After expanding all states,
we prune available beam states with lower path scores.
For example, we prune states 1-4, 3-4 and 3-2 in step2 of Figure \ref{fig:beam-softnms}.
In addition, as our needed output is order-independent,
we also prune duplicate states, e.g., the state 3-1 (equivalent to the state 1-3) for the diversity of beam states.
When all states in the beam can not be expanded or have been expanded before but failed,
we select the beam with the largest path score as the final output.

\section{Experiments}
\subsection{Experiments Setup}
\paragraph{Datasets}
We evaluate \modelname on FewNERD \cite{FewNERD} and SNIPS \cite{snips}.
\textbf{FewNERD} 
designs an annotation schema of $8$ coarse-grained (e.g., `Person') entity types and $66$ fine-grained (e.g., `Person-Artist') entity types, and constructs two tasks. One is \textbf{FewNERD-INTRA}, where all the entities in the training set (source domain), validation set and test set (target domain) belong to different coarse-grained types.
The other is \textbf{FewNERD-INTER}, where only the fine-grained entity types are mutually disjoint in different sets.
For the sake of sampling diversity, 
FewNERD adopts the $N$-way $K\sim2K$-shot sampling method to construct tasks (each class in the support set has $K\sim2K$ annotated entities).
Both FewNERD-INTRA and FewNERD-INTER have $4$ settings, $5$-way $1\sim2$-shot,  $5$-way $5\sim10$-shot,  $10$-way $1\sim2$-shot and  $10$-way $5\sim10$-shot. \textbf{SNIPS} is a slot filling dataset, which contains $7$ domains $\mathcal{D} = \{\mathcal{D}_1, \mathcal{D}_2, ..., \mathcal{D}_7\}$.
\cite{fewshothou} constructs few-shot slot filling task with the leave-one-out strategy, which means when testing on the target domain $\mathcal{D}_i$, they randomly chose $\mathcal{D}_j (i \ne j)$ as the validation domain, and train the model on source domains $\mathcal{D} - \{\mathcal{D}_i, \mathcal{D}_j \}$.
In the sampling task of SNIPS, all classes have $K$ annotated examples in the support set, but the number of them ($N$) is not fixed.
The few-shot slot filling task in each domain of SNIPS has two settings, $1$-shot and $5$-shot.
FSSL models are trained and evaluated on tasks sampled from the source and target domain, respectively.
To ensure the fairness,
we use the public sampled data provided by \cite{FewNERD} for FewNERD\footnote{The FewNERD data we used is from \url{https://cloud.tsinghua.edu.cn/f/0e38bd108d7b49808cc4/?dl=1}, which corresponds to the results reported in \url{https://arxiv.org/pdf/2105.07464v6.pdf}.} and data provided by \cite{fewshothou} for SNIPS, to train and evaluate our model.

\paragraph{Parameter Settings}
Following previous methods \cite{fewshothou, FewNERD}, we use uncased BERT-base as our encoder.
We use Adam \cite{adam} as our optimizer.
We set the dropout ratio \cite{dropout} to $0.1$. 
The maximum span length $L$ is set to $8$.
For BSNMS, the beam size $b$ is 5.
Because FewNERD and SNIPS do not have nested instances, we set the threshold to filter false positive spans $\delta$ to $0.1$,  the threshold to decay span scores $k$ to $1e$-$5$ and the decay ratio $u$ to $1e$-$5$ to force the refining results have no nested spans.
More details of our parameter settings are provided in Appendix \ref{app:exp}. 
\paragraph{Evaluation Metrics}
For FewNERD, following \cite{FewNERD}, we report the micro F1 over all test tasks, and the average result of $5$ different runs.
For SNIPS, following \cite{fewshothou}, we first calculate micro F1 score for each test episode (an episode contains a number of test tasks), and then report the average F1 score for all test episodes as the final result.
We report the average result of $10$ different runs the same as \cite{fewshothou}.

\paragraph{Baselines}
For systematic comparisons, we introduce a variety of baselines, including
\textbf{ProtoBERT} \cite{FewNERD,fewshothou}, \textbf{NNShot} \cite{FewNERD}, \textbf{StructShot} \cite{FewNERD},
\textbf{TransferBERT} \cite{fewshothou}, \textbf{MN+BERT} \cite{fewshothou}, \textbf{L-TapNet+CDT} \cite{fewshothou},
\textbf{Retriever} \cite{span-naacl}, \textbf{ConVEx} \cite{convex} and \textbf{Ma2021} \cite{mrc-few-slot}.
Please refer to the Appendix \ref{app:exp} for more details.

\subsection{Main Results}

Table \ref{tab:fewnerd} illustrates the results on FewNERD. 
As is shown, among all task settings, \modelname consistently outperforms ProtoBERT, NNShot and StructShot  by a large margin.
For example, compared with StructShot, \modelname achieves $11.17$ and $10.60$  average F1 improvement on INTRA and INTER, respectively.
Table \ref{tab:snips} shows the results on SNIPS.
In the $1$-shot setting, L-TapNet+CDT is the previous best method.
Compared with L-TapNet+CDT, \modelname achieves comparable results and $4.58$ F1-scores improvement in the $1$-shot and $5$-shot settings, respectively.
We think the reason is that the information benefits brought by our cross span attention mechanism in $1$-shot setting is much less than that in $5$-shot setting.
In addition, compared with L-TapNet+CDT, \modelname performs more stable, and also has a better model efficiency (Please refer to Section \ref{sec:me}).
In $5$-shot setting, Ma2021 is previous best method, and \modelname outperforms it $1.14$ and $0.42$ F1-scores in $1$-shot and $5$-shot settings, respectively.
These results prove the effectiveness of \modelname in few-shot sequence labeling.
\footnote{Results on SNIPS are worse than that reported in our first version paper \url{https://arxiv.org/pdf/2109.13023v1.pdf}, since we fix a data processing bug.}.

\begin{table}[t]
\centering
\small
\scalebox{1}{
    \begin{tabular}{lc}
    \toprule
    \textbf{Ablation Models} & \textbf{F1}   \\
    \midrule
    \modelname   & 61.7\tiny{$\pm$1.3}   \\
    \midrule

    \textit{r.m.} intra span attention (ISA) & 61.0\tiny{$\pm$0.4}  \\
    \textit{r.m.} cross span attention (CSA) & 59.1\tiny{$\pm$1.1}  \\
    \midrule
    \textit{r.m.} instance span attention (INSA) & 58.2\tiny{$\pm$1.3}  \\
    \textit{r.m.} O partition (OP) & 60.6\tiny{$\pm$1.3}  \\
    \midrule
    \textit{r.m.} beam soft-nms (BSNMS) & 57.3\tiny{$\pm$1.4}  \\
    \bottomrule
    \end{tabular}
}
\caption{The effect of our proposed mechanisms on the validation set of SNIPS (1-shot, domain ``We''). We report the average result of 3 different runs with standard deviations. \textit{r.m.} denotes \textit{remove}.}
\label{tab:abla-study}
\end{table}

\section{Analysis}
\subsection{Ablation Study}

To illustrate the effect of our proposed mechanisms,
we conduct ablation studies by removing one component of \modelname at a time.
Table \ref{tab:abla-study} shows the results on the validation set of SNIPS ($1$-shot, domain ``We'').
Firstly, we remove ISA or CSA, which means the span cannot be aware of other spans within the same sentence or spans from other sentences.
As shown in Table \ref{tab:abla-study}, 
the average F1 scores drop $0.7$ and $2.6$ without ISA and CSA, respectively.
These results suggest that our span enhancing module with ISA and CSA is effective.
In addition, CSA is more effective than ISA, since CSA can enhance query spans with whole support set spans, while ISA enhances spans only with other spans in the same setence.
CSA brings much more information than ISA.
Secondly, when we remove INSA and achieve the prototypical representation of a class through an average operation, the average F1 score would drop $3.5$.
When we do not consider the sub-classes of O-type spans (\textit{r.m.} OP), the average F1 score would drop $1.1$.
These results show that our span prototypical module is necessary.
At last, the result without BSNMS suggests the importance of our post-processing algorithm in this span-level few-shot labeling framework. Moreover, with BSNMS, \modelname can also easily adapt to the nested situation.

\subsection{Robustness in the Nested Situation}
\begin{figure}[t]
    \centering
    \includegraphics[width=0.75\linewidth]{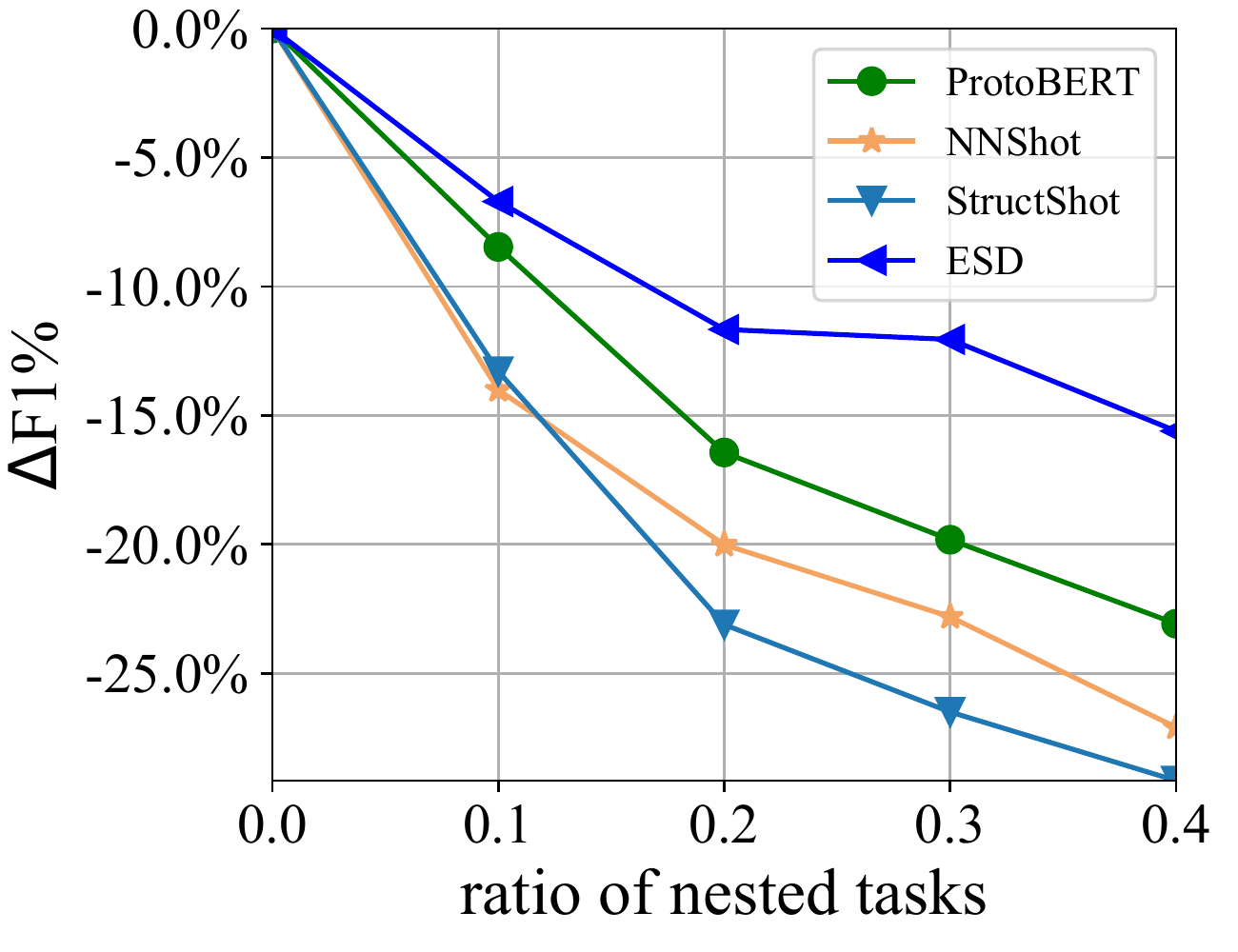}
    \caption{The performance drops in nested situations when increasing $r_{nest}$. $\Delta$F1\% is the F1 reduction percent over $r_{nested}=0$.
    }
    \label{fig:nested}
\end{figure}

Sequence labeling tasks such as NER can have nested situations. 
For example, in the sentence ``Isaac Newton studied at Cambridge University", where ``Cambridge'' is a location and ``Cambridge University'' is an organization.
However, both FewNERD and SNIPS do not annotate nested examples.
To explore the robustness of \modelname on such a challenging but common situation,  we construct \textbf{FewNERD-nested}, which has the same training tasks as FewNERD-INTRA, but different test tasks with a nested ratio $r_{nested}$.
In FewNERD-nested, we sample each test task either from FewNERD or from GENIA \cite{GENIA} with the probability $1 - r_{nested}$ and $r_{nested}$, respectively, and all tasks sampled from GENIA are guaranteed to have the query with nested entities.
We sample validation tasks with nested instances from ACE05 \cite{Ace05} to tune $k$, $u$ and $\delta$ in BSNMS. 
Please refer to the Appendix \ref{app:nest} for more details about FewNERD-nested.
Figure \ref{fig:nested} shows the results of \modelname and several typical baselines in FewNERD-nested with different $r_{nested}$.
When $r_{nested}$ increases, \modelname is more stable than previous methods, since \modelname with BSNMS can easily extend to nested tagging cases without any extra training while previous methods are incapable of.
In addition, we also compare different post-processing methods when $r_{nested}=1$.
As shown in Table \ref{tab:post}, 
the beam search method can not handle the nested situation, and thus it harms the performance.
SoftNMS \cite{SoftNMS-NER} and BSNMS can both improve the model performance.
However, when incorporating beam search into SoftNMS, the model can be more flexible and avoid some local optimal post-processing results achieved by SoftNMS (e.g., spans $1$,$3$ and $4$ in Figure 3), and thus BSNMS outperforms SoftNMS \footnote{\modelname is also more robust than baselines in the noisy situation. Please refer to Appendix \ref{sec:noisy} for details.}.

\begin{table}[t]
\centering
\scalebox{0.9}{
    \begin{tabular}{lc}
    \toprule
    \textbf{Models} & \textbf{F1}   \\
    \midrule
    \modelname (with BSNMS)    & 31.25   \\
    \midrule
    \modelname (with SoftNMS) & 31.09 \\
    \modelname (with Beam Search) & 28.94 \\
    \modelname (without Post-processing) & 30.34 \\
    \bottomrule
    \end{tabular}
}
\caption{Comparison between different post-processing methods in the nested situation with $r_{nested}=1$.}
\label{tab:post}
\end{table}

\subsection{Model Efficiency}
\label{sec:me}

\begin{table}[t]
\small
\centering
\begin{tabular}{lccc}
\toprule
\multirow{2}{*}{\textbf{Models}} & \multirow{2}{*}{\textbf{\#Para.}} & \multicolumn{2}{c}{\textbf{Inference Time}} \\ 
\cmidrule(r){3-4} 
~            & ~      & \textbf{\textsc{1-shot}}    & \textbf{\textsc{5-shot}}        \\    
\midrule
\modelname   & 112M      & 8.53 ms   & 18.47 ms         \\
\midrule
ProtoBERT & 110M & 3.13 ms & 5.27 ms \\
L-TapNet+CDT & 110M      & 24.67 ms   & 54.19 ms         \\ 
\bottomrule
\end{tabular}
\caption{The parameter number and inference time per task of \modelname and L-TapNet+CDT in the domain ``We'' of SNIPS. Although \modelname is slower than ProtoBERT,
\modelname outperforms ProtoBERT by $31.53$ and $16.68$ F1 scores in 1- and 5-shot settings with a bearable latency.}
\label{tab:comp}
\end{table}

Compared with the token-level models, \modelname needs to enumerate all spans within the length $L$ for a sentence.
Therefore, the number of spans is approximately $L$ times that of tokens, which may bring extra computation overhead.
To evaluate the efficiency of \modelname, we compare the average inference time per task of \modelname (including the BSNMS post-processing process), L-TapNet+CDT (the state-of-the-art token-level baseline model with the open codebase in the SNPIS) and ProtoBERT (an extremely simple token-level baseline).
As shown in Table \ref{tab:comp}, with exactly the same hardware setting,
in the domain ``We'' of SNIPS 1-shot setting, \modelname (avg. $8.53$ ms per task) is nearly 3 times faster than L-TapNet+CDT (avg. $24.67$ ms per task). We see a similar tendency in the 5-shot setting.
Although \modelname (avg. $8.53$  and $18.47$ ms in 1- and 5-shot per task) is slower than ProtoBERT (avg. $3.13$ and $5.27$ ms),
it outperforms ProtoBERT by $31.53$ and $16.68$ F1 scores in 1- and 5-shot settings (reported in Table \ref{tab:snips}) with a bearable latency.
Besides the inference time, we also compare the parameter number of these models.
As is shown, the added parameter scale (span enhancing and prototypical modules) is very small (2M) compared with the ProtoBERT and L-TapNet+CDT (110M). 
These results show that \modelname has an acceptable efficiency.

\begin{table}[t]
\centering
\scalebox{0.86}{
\begin{tabular}{lccccc} \toprule
\textbf{Models} & \textbf{F1} & \textbf{Total} & \textbf{FP-Span} & \textbf{FP-Type}  \\
\midrule
\modelname  & 59.29 &     9.4k     &   72.8\%   &   27.2\%  \\  
\midrule
ProtoBERT & 38.83 & 30.4k   &   86.7\%  &   13.3\%   \\
NNShot   &  47.24 & 21.7k    &  84.7\%  &   15.3\%  \\
StructShot & 51.88 & 14.5k    &   80.0\%  &   20.0\%   \\ 
\bottomrule  
\end{tabular}}
\caption{Error analysis of $5$ way $1\sim2$ shot on FewNERD-INTER. `FP-Span' denotes extracted entities with the wrong span boundary, and `FP-Type' represents extracted entities with the right span boundary but the wrong entity type. `Total' denotes the total wrong prediction of two types.}
\label{tab:error}
\end{table}

\subsection{Error Analysis}
To  further explore  what  types  of  errors the model makes in detail, we divide error of model prediction into $2$ categories, `FP-Span' and `FP-Type'.
As shown in Table \ref{tab:error}, 
\modelname outperforms baselines and has much less false positive prediction errors.
`FP-Span' is the major prediction error for all models, showing that it is hard to locate the right span boundary in FSSL for existing models.
Therefore, we should pay more attention to the span recognition in the future work.
However, compared with previous methods, \modelname has less ratio of the `FP-span' error.
We think the reason is that our span-level matching framework with a series of span-related procedures has a better perception of the entity and slot spans than that of our baselines.

\section{Conclusion}
In this paper, we propose \modelname, an enhanced span-based decomposition model for few-shot sequence labeling (FSSL).
To overcome the drawbacks of previous token-level methods,
\modelname formulates FSSL as a span-level matching problem, 
and decomposes it into a series of span-related procedures, mainly including span representation, class prototype aggregation and span conflicts resolution for a better span matching.
Extensive experiments show that \modelname achieves the state-of-the-art performance on two popular few-shot sequence labeling benchmarks and that ESD is more robust than previous models in the noisy and nested situation.

\section*{Acknowledgements}
The authors would like to thank the anonymous reviewers for their thoughtful and constructive comments.
This paper is supported by the National Key Research and Development Program of China under Grant No. 2020AAA0106700, the National Science Foundation of China under Grant No.61936012 and 61876004, and NSFC project U19A2065.

\bibliography{FSSL}
\bibliographystyle{acl_natbib}

\clearpage
\appendix
\section{Experiments}
\label{app:exp}
\subsection{Baselines}
We compare \modelname with a variety of baselines as follows:
\begin{itemize}
    \item \textbf{TransferBERT} \cite{fewshothou} is a fine-tuning based model, which is a direct application of BERT \cite{Bert} to the few-shot sequence labeling.
    \item \textbf{ConVEx} \cite{convex} is a fine-tuning based model, which is first pre-trained on the Reddit corpus with the sequence labeling objective tasks and then fine-tuned on the source domain and target domain annotated data for final few shot sequence labelingm    \item \textbf{Ma2021} \cite{mrc-few-slot} formulates sequence labeling as the machine reading comprehension problem, and proposes some questions to extract slots in the query sentence.
    \item \textbf{ProtoBERT} \cite{WarmProtoZero}  predicts the query labels according to the similarity of BERT hidden states of support set and query tokens.
    \item  \textbf{Matching Net (MN)+BERT} \cite{fewshothou} is similar to ProtoBERT. The only difference is that MN uses the matching network \cite{MN} for token classification.
    \item \textbf{L-TapNet-CDT} \cite{fewshothou} utilizes the task-adaptive projection network \cite{yoon2019tapnet}, pair-wise embedding and collapsed dependency transfer mechanisms to do classification.
    \item  \textbf{NNShot} \cite{NNshot} is similar to ProtoBERT, while it makes the prediction based on the nearest neighbor.
    \item  \textbf{StructShot} \cite{NNshot} adopts an additional Viterbi decoder during the inference phase on top of NNShot. 
    \item \textbf{Retriever} \cite{span-naacl} is a retrieval based method which does classification according to the most similar example in the support set.
\end{itemize}

\subsection{Parameter Setting}
In our implementation, 
we utilize \textit{BERT-base-uncased} as our backbone encoder the same as \cite{fewshothou, span-naacl, FewNERD}. 
We use Adam \cite{adam} as our optimizer.
In FewNERD, the learning rate is $2e-5$ for BERT encoder and $5e-4$ for other modules.
In the 1-shot setting of SNIPS, for the domain ``Mu'', the learning rate is $5e-6$ for BERT encoder and $1e-4$ for other modules.
For the domain ``Bo'', the learning rate is $1e-5$ for BERT encoder and $1e-4$ for other modules.
For other settings of SNIPS, the learning rate is $5e-5$ for BERT encoder and $5e-4$ for other modules.
We set the dropout ratio \cite{dropout} to $0.1$. 
The dimension of span representation $d$ and the maximum span length $L$ is set to $100$ and $8$, respectively.
For BSNMS, the beam size $b$ is 5.
Since these is no nested instances in FewNERD and SNIPS, we set  the threshold to filter false positive spans $\delta$ to $0.1$,  the threshold to decay span scores $k$ to $1e-5$ and the decay ratio $u$ to $1e-5$ to force the refining results have no nested spans.
We use the grid search to search our hyperparameters, and the scope of each hyperparameter are included in Table \ref{tab:hyperparameters}.
We train our model on a single NVIDIA A40 GPU with 48GB memory. 
\begin{table}[h]
    \centering
    \scalebox{0.9}{
    \begin{tabular}{lc}
    \toprule
       learning rate  & [5e-5, 1e-4, 3e-4, 5e-4] \\
       dropout & [0.1, 0.3, 0.5] \\
       bert learning rate & [5e-6, 1e-5, 2e-5, 3e-5, 5e-5] \\
       span dimension & [50, 100, 150, 200] \\
       beam size & [1, 3, 5, 7] \\
    \bottomrule 
    \end{tabular}
    }
    \caption{The searching scope of hyperparameters.}
    \label{tab:hyperparameters}
\end{table}

\section{FewNERD-nested}
\begin{table}[h]
    \centering
    \scalebox{0.9}{
    \begin{tabular}{lc}
    \toprule
         $k$ & [0.1, 0.2, 0.3, 0.4, 0.5, 0.6, 0.7, 0.8, 0.9] \\
       $\delta$ & [0.1, 0.2, 0.3, 0.4, 0.5, 0.6, 0.7, 0.8, 0.9] \\
       $u$ & [0.1, 0.2, 0.3, 0.4, 0.5, 0.6, 0.7, 0.8, 0.9] \\
    \bottomrule 
    \end{tabular}
    }
    \caption{The searching scope of BSNMS hyperparameters.}
    \label{tab:nest-hyperparameters}
\end{table}
\label{app:nest}
We construct our FewNERD-nested via ACE05 \cite{Ace05}, GENIA \cite{GENIA} and the origin FewNERD datasets.
GENIA is a biological named entity recognition dataset, which contains 5 kinds of entities, \textit{`DNA'}, \textit{`Protein'}, \textit{`cell\_type'}, \textit{`RNA'} and \textit{`cell\_line'}, and
all of these entity types are not included in the FewNERD.
We partition the sentences in GENIA into two groups, $\mathcal{G}_1$ and $\mathcal{G}_2$.
$\mathcal{G}_1$ consists of sentences with nested entities ($4,744$ sentences in total), and $\mathcal{G}_2$ consists of sentences without nested entities ($11,924$ sentences in total).
We utilize sentences in $\mathcal{G}_1$ and $\mathcal{G}_2$ to construct the query and support set of the GENIA task, respectively.
Our FewNERD-nested contains 2000 5-way 5$\sim$10-shot test tasks in total, where
$r_{nested}$ percent tasks are from GENIA, and the remained tasks are from FewNERD-INTRA.
We use ACE05 to construct the validation set.
ACE05 is a widely used named entity recognition dataset, which contains $7$ coarse-grained entity types, \textit{`FAC'}, \textit{`PER'}, \textit{`LOC'}, \textit{`VEH'}, \textit{`GPE'}, \textit{`GPE'}, \textit{`WEA'} and \textit{`ORG'}.
The \textit{`LOC'}, \textit{`PER'}, \textit{`ORG'} and \textit{`FAC'} are not included in the training set of FewNERD-INTRA, and therefore we use them ($28$ fine-grained entity types in total) and sample $1000$ nested tasks as the validation dataset to tune the $k$, $\delta$ and $u$ of BSNMS.
The search scope is included in the Table \ref{tab:nest-hyperparameters}.
In this nested situation, we finally set the $k$, $\delta$ and $u$ to $0.1$, $0.1$ and $0.4$ respectively.

\section{Robustness in the Noisy Situation}
\begin{figure}[t]
    \centering
    \includegraphics[width=0.75\linewidth]{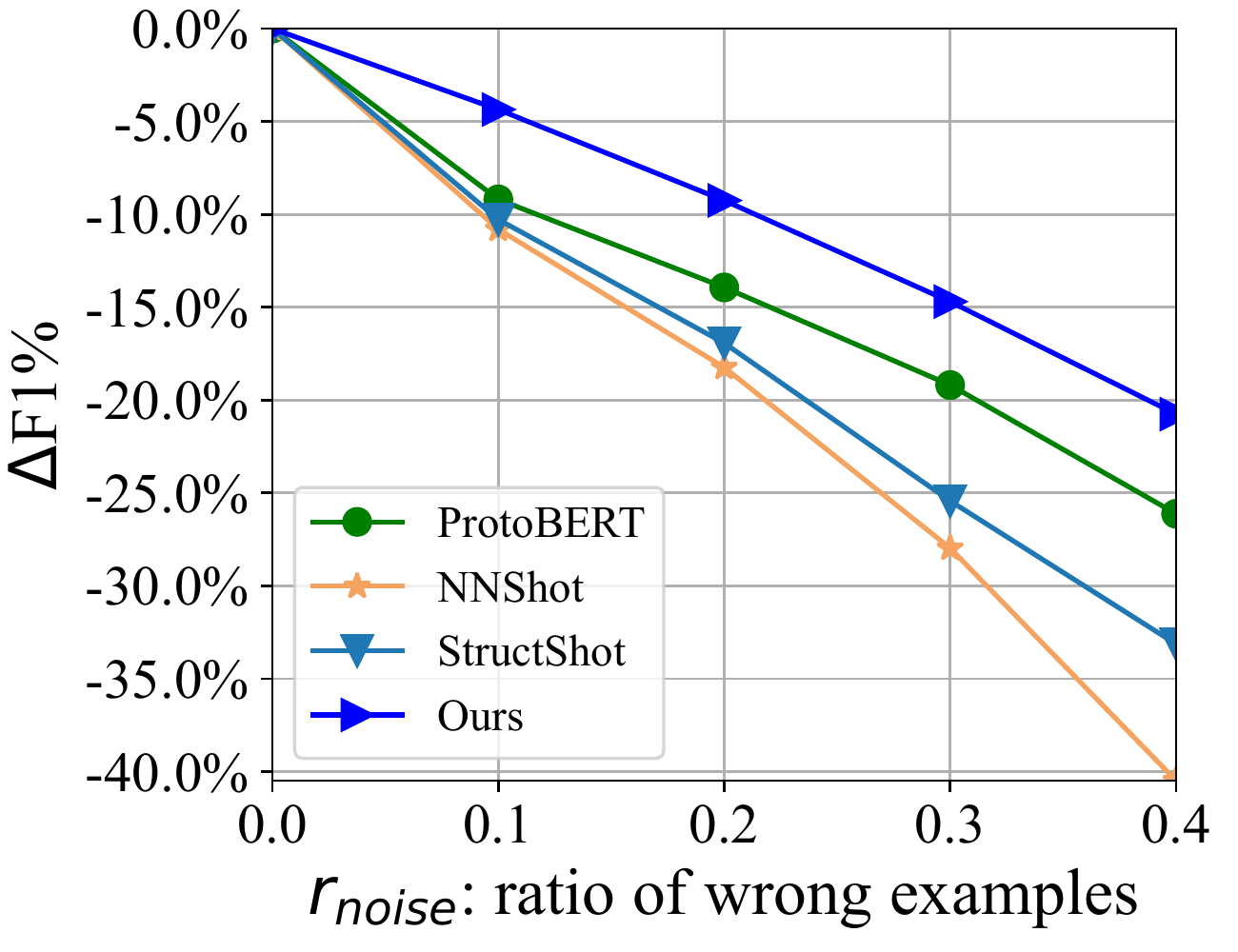}
    \caption{The performance drops in noisy situations when increasing $r_{noise}$. $\Delta$F1\% is the F1 reduction percent over $r_{noise}=0$.
    }
    \label{fig:noise}
\end{figure}
\label{sec:noisy}
FSSL methods tend to be seriously influenced by the noise in the support set, since they make the prediction based on only limited annotated examples.
To explore the robustness of \modelname in the noisy situation, we construct \textbf{FewNERD-noise}.
In FewNERD-noise, we disturb each  $5$ way $5\sim10$ shot FewNERD-INTRA task with a noisy ratio $r_{noise}$, which means there are nearly $r_{noise}$ percent entities in the support set are mislabeled.
As illustrated in the right part of Figure \ref{fig:noise}, with $r_{noise}$ increasing, the performance of \modelname drops less than baselines, which furthuer shows the superiority of our methods.

\section{A Detail Case Study of BSNMS}
\begin{figure*}[h]
    \centering
    \includegraphics[width=\linewidth]{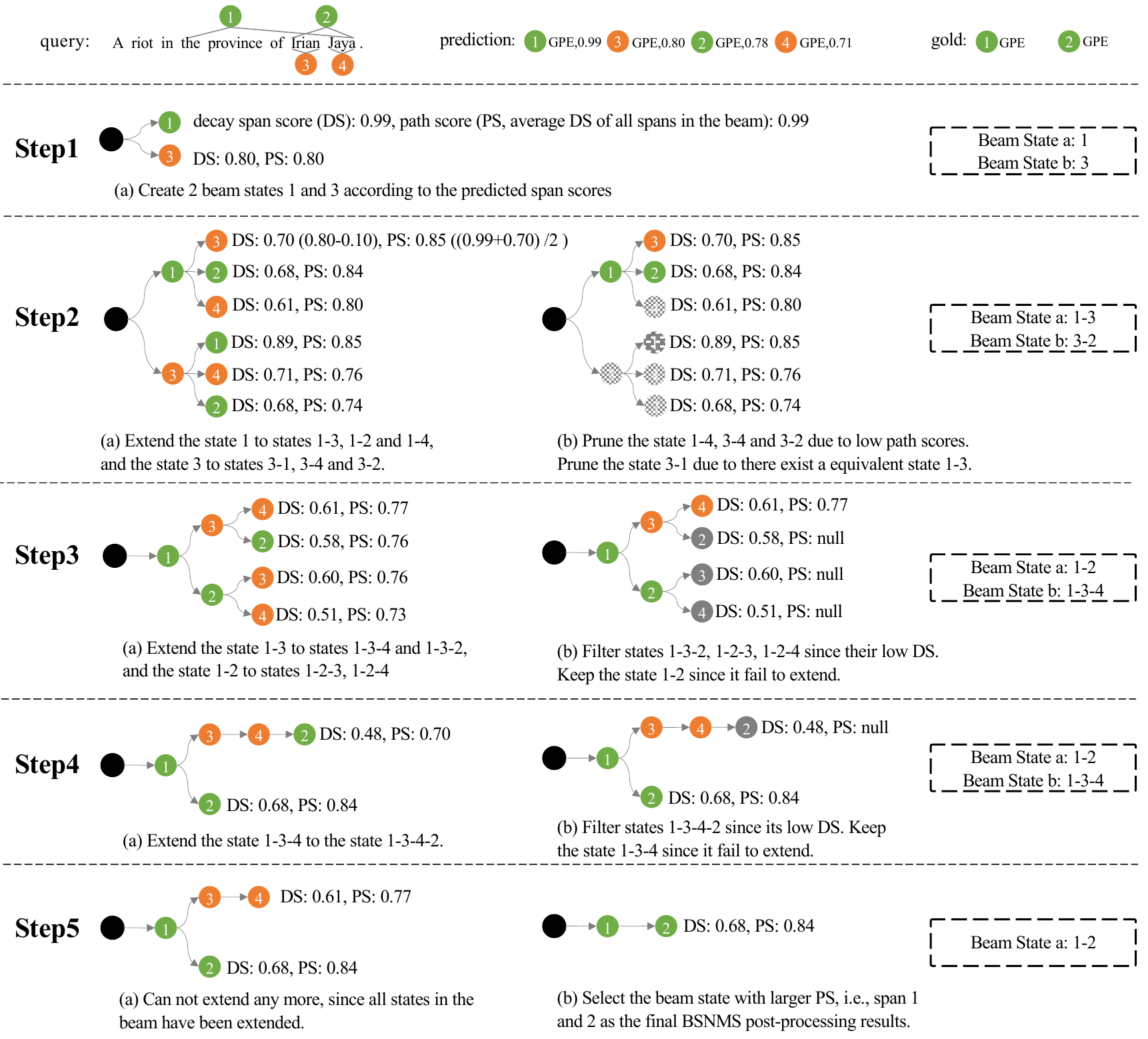}
    \caption{A step by step processing process of BSNMS with beam size=2. For clarity, we set the filter threshold $\delta$ to $0.6$, and suppose the span score is always decayed by the overlapped spans with a constant decayed score  $-0.1$. 
    \textbf{STEP1}:create $2$ (beam size) states with spans having larger predicted scores; 
    \textbf{STEP2}: extend all states (i.e., states 1 and 3) in the beam to 1-3, 1-2, 1-4, 3-1, 3-4 and 3-2. Compute DS of the new added span and PS of the new state. As beam size=2, prune states 1-4, 3-4 and 3-2 according to their lower PS. The state 3-1 is dropped since  it is equivalent with the state 1-3;
    \textbf{STEP3}: extend states in the beam (1-3 and 1-2) to 1-3-4, 1-3-2, 1-2-3, 1-2-4, and compute their DS and PS. Filter states 1-3-2, 1-2-3, 1-2-4 since their DS are not greater than $\delta$, i.e., low DS;
    \textbf{STEP4}: extend states in the beam (only 1-3-4 since the state 1-2 has been extended before) to 1-3-4-2, and filter 1-3-4-2 due to its low DS;
    \textbf{STEP5}: all states in the beam can not extend any more, and select the final state (1-2 in this case) with the largest PS.
    } 
    \label{fig:bsnms-detail}
\end{figure*}

For span conflicts resolution, we propose a post-processing method BSNMS.
A step by step demonstration of BSNMS is shown in Figure \ref{fig:bsnms-detail}.
\label{apx:bsnms}

\end{document}